\documentclass[10pt, twocolumn]{article}
\usepackage[backend=biber]{biblatex}
\usepackage{amsmath,amsfonts}
\usepackage{mathtools}
\usepackage{algorithmic}
\usepackage{algorithm}
\usepackage{array}
\usepackage[caption=false,font=normalsize,labelfont=sf,textfont=sf]{subfig}
\usepackage{textcomp}
\usepackage{stfloats}
\usepackage{url}
\usepackage{verbatim}
\usepackage{graphicx}
\usepackage{multicol}
\usepackage{siunitx}
\usepackage{booktabs}
\usepackage[margin=0.5in]{geometry}

\providecommand{\keywords}[1]
{
  \small	
  \textbf{\textit{Keywords---}} #1
}

\renewenvironment{abstract}
 {\small
  \begin{center}
  \bfseries  \section*{\abstractname} \vspace{-.5em}\vspace{0pt}
  \end{center}
  \list{}{%
    \setlength{\leftmargin}{0mm}
    \setlength{\rightmargin}{\leftmargin}%
  }%
  \item\relax}
 {\endlist}

\begin{document}

\title{Deep Learning-Based Image Recovery and Pose Estimation for Resident Space Objects}

\author{Louis Aberdeen, Mark Hansen, Melvyn L. Smith, Lyndon Smith\\%
 University of the West of England, Centre for Machine Vision,\\ Coldharbour Ln, Stoke Gifford, Bristol BS16 1QY.
     \thanks{This work was supported by Metrea Mission Data, the University of the West of England and Innovate UK. Funded as part of a Knowledge Transfer Partnership. KTP reference 13810.}   %
}

\maketitle

\begin{abstract}
As the density of spacecraft in Earth’s orbit increases, their recognition, pose and trajectory identification becomes crucial for averting potential collisions and executing debris removal operations. However, training models able to identify a spacecraft and its pose presents a significant challenge due to a lack of available image data for model training. This paper puts forth an innovative framework for generating realistic synthetic datasets of Resident Space Object (RSO) imagery, Using the International Space Station (ISS) as a test case, it goes on to combine image regression with image restoration methodologies to estimate pose from blurred images. An analysis of the proposed image recovery and regression techniques was undertaken, providing insights into the performance, potential enhancements and limitations when applied to real imagery of RSOs. The image recovery approach investigated involves first applying image deconvolution using an effective point spread function, followed by detail object extraction with a U-Net. Interestingly, using only U-Net for image reconstruction the best pose performance was attained, reducing the average Mean Squared Error in image recovery by 97.28\% and the average angular error by 71.9\%. The successful application of U-Net image restoration combined with the Resnet50 regression network for pose estimation of the International Space Station demonstrates the value of a diverse set of evaluation tools for effective solutions to real-world problems such as the analysis of distant objects in Earth’s orbit.
\end{abstract}


\section*{Introduction}
\par The quantity of space debris and close approaches of satellites has been steadily increasing in recent years, posing significant challenges to the continued safe and reliable operation of space-based assets. As more countries and private companies launch satellites into orbit, the number of objects orbiting the Earth has appreciably increased, The European Space Agency estimates there are roughly 34,000 objects larger than 10 cm \cite{ESAdebris} in orbit. This debris, which includes defunct satellites, rocket bodies, and fragments from past collisions and explosions, can travel at speeds of up to 7-8 kilometres per second, creating a hazardous environment for active spacecraft. This trend is expected to continue as the demand for space-based services, such as communications, navigation, and Earth observation, continues to increase globally. Even though space missions are currently supported by the large U.S. Space Surveillance Network, who track space debris, most have noticed the necessity of creating multiple streams of data. By integrating data from various sensors, satellites, and ground-based systems, space mission planners can make more informed decisions, reduce mission risks, and optimize resource allocation. This approach ensures that all aspects of the mission are considered, leading to safer and more successful space operations.

\par The demand for Space Domain Awareness (SDA) capabilities has grown significantly in recent years. However, current methods and technologies used for tracking and monitoring space debris have limitations in their ability to provide comprehensive and timely information about the evolving space environment. For instance, ground-based electro-optical systems are often limited by weather and atmospheric conditions, affecting their ability to capture actionable data. Similarly, radar systems have limited coverage and face challenges in characterizing objects. In-orbit solutions can capture extremely high-quality images of resident space objects, however, the cost to deploy and maintain such systems is prohibitively high. To address these challenges, new and innovative approaches to space domain awareness are needed.

\par This is where the use of advanced sensor technologies and data analytics can play a crucial role in monitoring the space environment effectively.

This paper addresses the challenge of enhancing imagery collected by the commercial organization Metrea Ltd using its proprietary technology. We focus on improving the quality of these images to infer details about the objects of interest, specifically the International Space Station (ISS). Our goal is to determine the station’s pose, while ensuring that the techniques we develop are as general as possible, allowing for expansion to other classes of RSOs.

Multiple challenges hinder capturing clear photos or videos of spacecraft from the ground:
\begin{itemize}
\item \textbf{Atmospheric disturbances.} The Earth’s atmosphere causes light to be disturbed, creating visual distortions, or atmospheric turbulence. This blurs the images observed from the ground.
\item \textbf{Distance.} Many spacecraft fly at altitudes of thousands of kilometres. Even high-magnification telescopes have difficulty capturing good-quality images of these small distant targets.
\item \textbf{Speed.} RSOs (Resident space objects) in low Earth orbit usually travel at about 7-8 kilometres per second. This makes it difficult to track and image them.
\item \textbf{Light conditions.} Depending on the spacecraft's location, the Sun's position and the observation site, different light conditions may occur. This may cause the spacecraft to be backlit or receive too little light, increasing the difficulty of good imaging.
\item  \textbf{Equipment} Taking high-quality photographs of objects in space requires equipment with highly advanced specifications/capabilities. Where technical requirements may be technologically unavailable or prohibitively expensive.

\end{itemize}

As the demand for increasingly accurate object identification grows, it has become imperative to explore the potential of deep learning and machine vision techniques \cite{SMITH2021103472}. 

To develop a viable solution, the following requirements must be considered:

\begin{itemize}
    \item \textbf{Light weight.} The model and processing should take only a few seconds on a conventional GPU-equipped laptop, ensuring quick display. Being able to discern features quickly is critical for making informed and timely decisions.
\item \textbf{Accurate.} The system must create images that accurately represent the object as closely as possible. This is essential for the attribution and classification of RSOs.
\item  \textbf{General.} The methods should not be limited to apply to one camera system or just one type of object. This will ensure generalization.
\end{itemize}

\par The main obstacle to building any deep learning system is having a sufficiently large and diverse dataset from which the model can learn. This is certainly an issue for this application, where we do not have the high-quality target images to build a traditional training loop. Therefore, we have chosen to generate our exemplar images synthetically using a physics-based renderer \cite{black2021realtimeflightreadynoncooperativespacecraft} and to simulate the degradation processes observed in real images. This approach allows us to train our model to learn the inverse transformation needed to recover the identity and pose of unknown objects from real images. 

\section*{Literature Review}
\par The backbone of most image-related tasks, CNNs have become the de facto standard for image regression \cite{ZHANG2022110589, 8970273, pmlr-v121-zhang20a}. Leveraging pre-trained models, primarily designed for classification tasks like ResNet \cite{he2015deepresiduallearningimage}, VGG \cite{simonyan2015deepconvolutionalnetworkslargescale}, and EfficientNet\cite{tan2020efficientnetrethinkingmodelscaling} has been a prevalent approach. By adapting the final classification layer to a regression output, these models can harness the rich feature representations learned from datasets like ImageNet \cite{5206848}.

Images of RSOs are likely to be blurred and of low resolution, so attempting to recover more details reliably seems a good first step.

One approach to image recovery is Super-Resolution where we teach a network to interpolate the pixel values when upscaling an image. Many works have developed image restoration methods based on GAN \cite{goodfellow2014generativeadversarialnetworks}, such as SRGAN \cite{ledig2017photorealisticsingleimagesuperresolution}, ESRGAN \cite{wang2018esrganenhancedsuperresolutiongenerative}, and Real-ESRGAN \cite{wang2021realesrgantrainingrealworldblind}. SRGAN and ESRGAN introduce perceptual loss to enhance the visual quality of images. Real-ESRGAN employs a downsample network to learn the image degradation in the real world, aiming to generate synthetic low-resolution images that are more suitable for model training. However, despite its advantages, GAN still suffers from issues like mode collapse, difficulty in convergence, and the potential for generators to take erroneous shortcuts \cite{saxena2023generativeadversarialnetworksgans}.

In \cite{sharma2019poseestimationnoncooperativerendezvous} the authors introduce the Satellite Pose Network for known non-cooperative spacecraft using a single gray scale image from monocular vision. A Convolutional Neural Network (CNN) is used with three branches: one for placing a 2D bounding box around the spacecraft, the second branch classifies the input region into coarse attitude labels, and then the third branch refines the estimate to a finer attitude. The method also introduces the Spacecraft Pose Estimation Dataset (SPEED) \cite{Kisantal_2020} for training and evaluation, which includes synthetic and actual images of the Tango spacecraft. They develop on the SPEED dataset with the SPEED+ dataset \cite{Park_2022} by including hardware-in-the-loop images of a spacecraft mock-up model captured from a test bed they designed.

In \cite{Park_2022} a dataset is opened as a competition, with the best-performing models using Perspective-n-Point (PnP) solvers to match features in the image to vertices in a wire mesh model of the uncooperative satellite. This approach uses an iterative minimization scheme such as Gauss-Newton or Levenberg-Marquardt to solve the PnP problem and generalizes well \cite{Kisantal_2020}, even when unseen features like the Earth are present in the background of the image. However, this method relies on having both a wireframe model of the satellite and the features being distinct enough for accurate alignment.

In this paper, we use deep transfer learning on the task of pose estimation. Deep transfer learning involves taking a learned feature mapping from a source dataset or source domain (even one not strongly related to the dataset) and applying this to a target dataset/target domain. This is usually done to reduce learning costs and improve the generalizability of models for the target. In many ML problems, arranging a large amount of labelled data is very difficult. This approach allows pre-trained models to be applied to new target domains by fine-tuning the lateral layers of the network with what limited labelled data is available. Here we apply transductive transfer learning as we know the labels in our source domain but do not have labels in our target domain. \cite{zhuang2020comprehensivesurveytransferlearning,inproceedings} 

U-Net is a model originally developed for medical image segmentation \cite{ronneberger2015unetconvolutionalnetworksbiomedical}, however, it has since been applied in many other contexts. It has been shown to perform well in image recovery \cite{roy2019lunarsurfaceimagerestoration}, pan-sharpening \cite{YAO2018364} and image enhancement tasks \cite{chen2018learningdark}. The model uses a contracting path to extract feature maps at different spatial dimensions to enhance localization. Here we instead apply U-Net to the image recovery problem instead.

In the paper \cite{Akhaury_2022} the authors explore using deconvolution techniques such as Tikhonov regularization \cite{tikhonov} followed by U-Net, A wavelet \cite{Wavelets} and an x64 U-Net, a larger version of the original U-Net with ~31M trainable parameters. We similarly focus on the deconvolution + U-Net approach as it showed the most promise. This is because we had already learned the Point Spread Function (PSF) to make our dataset; additionally, deconvolution is a general approach that can be applied to any images captured using the optical system (not just images of the ISS). 

\section*{Motivation}
Image recovery and image enhancement are two related but different topics. Image recovery is used to estimate the original clean image given a noisy/corrupted version. Image enhancement is intended to make the image more informative by highlighting important features. Image enhancement is much more qualitative in its approach whereas image recovery has a specific criterion to minimise.

Image recovery is an inverse problem where we seek to determine the original image from a noisy and blurred observation. We use Tikhonov regularization to stabilize the solution by adding a regularization term to the least-squares problem thus improving the generalization ability of models as shown in Eq. \eqref{eq1}.

\begin{equation}
    \min_{x}\{||A x - b||^2 + \lambda^2||Lx||^2 \}
    \label{eq1}
\end{equation}

where  $A$  is the matrix of the system, $b$ is the observed data,  $\lambda $ is the regularization parameter, and $ L $ is the regularization matrix. We explore three regularization techniques: Arnoldi-Tikhonov, Hybrid Generalized Minimal Residual and Golub-Kahan-Tikhonov. In general, a projection method for computing an approximation is defined by the two conditions shown in Eq. \eqref{eq2}. 
\begin{equation}
    \mathbf{x}_d \in \mathcal{S}_d, \quad \mathbf{r}_d \coloneq \mathbf{b} - \mathbf{A}\mathbf{x}_d \perp \mathcal{C}_d
    \label{eq2}
\end{equation}
Where $\mathcal{S}_d$ and $\mathcal{C}_d$ are the approximation and constraint subspace, respectively.

When representing the convolution as a linear operator, if our image is $m$ by $n$, then we represent our image as a $mn$ by $1$ vector and our convolution as a $mn$ by $mn$ matrix, which in our case is $32,400$ by $32,400$. Standard Tikhonov regularization would require us to take a Single Value Decomposition (SVD) of our convolution operator. SVD by Jacobi rotations has a time complexity of $O(n^3)$ so is intractable for this and any larger problems. Because of this, we focus on projection methods to reduce the computational cost as follows:

\begin{itemize}
    \item \textbf{Arnoldi-Tikhonov} combines Arnoldi iteration with Tikhonov regularization. The approach projects the problem to a lower dimensional Krylov space using Arnoldi iteration where Tikhonov regularization is applied, and the solution is projected back to the full space.

    \item \textbf{Hybrid Generalized Minimal Residual} combines Generalised Minimal Residual steps and regularization steps. The GMRES selects the approximation and constraint subspaces such that the residual is minimized in the approximation subspace. The hybrid approach then applies iteration-dependent Tikhonov regularization to the projected problem.

    \item \textbf{Golub-Kahan-Tikhonov} combines the Golub–Kahan bidiagonalization algorithm with Tikhonov regularization. The approach projects the problem to a lower dimensional Krylov space in a fixed number of iterations then applies Tikhonov regularization and projects the solution to the full space.
\end{itemize}

Deconvolution can significantly enhance the clarity and sharpness of an image, making it easier to identify and analyze features that were previously obscured by blur and noise. Despite its advantages, deconvolution is sensitive to inaccuracies in the PSF and can amplify noise if not handled properly, so we investigate the effect deconvolution has on pose estimation and image recovery models for RSO imagery. Considering the non-linear image degradation effects in RSO imagery, like bloom, using a U-net architecture could be particularly apt because its encoder-decoder structure and skip connections effectively capture both global and local features, addressing complex distortions and enhancing the quality of image recovery.
\section*{Methodology}
\subsection*{Dataset generation}

We investigated how a limited dataset of real low quality optical imagery can be used to generate a much larger source training dataset. We then trained a model to enhance those images, compared the performance of a rotation estimation model trained with those datasets, and study the model performance when applied to actual imagery of the ISS.

The approach depends on generating a large artificial dataset that accurately replicates the imagery collected by the optical system employed. For our case study, we use a 3D model of the ISS and render it with Blender \cite{blender}, an open-source 3D rendering tool, to generate images. We then apply optical distortion to these images. To create high-quality synthetic images of the ISS, we utilize the Cycles engine, a physics-based production rendering engine that supports GPU acceleration. 

A solar illumination source intensity (1,360 watts per square meter) and a camera were employed to replicate authentic environmental conditions. This simulated camera captures images from various vantage points, specified by XYZ Euler angles, resulting in a set of 24x24x24=13,824 images at different orientations. Each image was annotated based on its corresponding Euler angle designation.

To produce accurate images as seen from the camera we simulate the propagation of the ISS's orbit using the Two Line Element (TLE) information, sourced from \textit{spacetrack.org} \cite{SpaceTrack} to get an accurate distance. However, due to sensitivity around the specifications of the telescope system used, we estimated the camera parameters to achieve a final image that is as similar as possible.

NASA's VTAD has produced a 3D model of the ISS \cite{ISSMODEL} that accurately replicates the ISS in both its shape and the reflectance/diffuse/specular material properties and this was used here.

\subsection*{Modeling optical degradation using the Point Spread Function}
RSO image emulators are typically bespoke and made for a particular optical system and mission \cite{s21237868} \cite{sharma2019poseestimationnoncooperativerendezvous}. Alternatively, an actual optical system is used in a physical test bed that emulates the conditions of RSO imagery \cite{Park_2022}. However, because the optical system used here does not have a publicly available schematic and was not extensively available for data collection, a different approach was needed which made use of available data.

We take inspiration from the Point-Spread-Function (PSF) \cite{PhysRevSTAB.1.062801} which describes a focused optical systems response to a single point source - used in astronomical imaging \cite{10.1117/12.892762}, Medical imaging \cite{Ashrafinia_2017}, and electron microscopy \cite{kamal2024pointspreadfunctionopticsscanning}.

The use of convolution for modelling image degradation in RSO imagery is based on principles of fundamental optical physics and linear systems theory. The imaging process can be mathematically represented as a linear, shift-invariant system (LSI) see Eq. \eqref{continuosconv}.
\begin{itemize}
    \item $I(m,n)$ is the observed image.
    \item $O(m,n)$ is the true image of the object.
    \item $PSF(x-x',y-y')$ is the point spread function.
\end{itemize}

\begin{align}
I(x, y) &= (O * PSF)(x, y) \label{continuosconv}\\
&= \int_{-\infty}^{\infty} \int_{-\infty}^{\infty} O(x', y') \cdot PSF(x - x', y - y') dx' dy'  
\end{align}

The point-spread function can encapsulate many degradation processes such as atmospheric turbulence, motion blur, and optical system effects. This formulation is applicable under the assumption that image degradation processes apply linearity and are shift-invariant which is generally the case when imaging RSOs. There can be significant variations temporally or when objects move through the frame in different directions. Ideally we would use stars collected either during or just before/after the object is sighted as these will best model the degradation seen in the imagery. In this case study we use stars collected with the same camera system at a different time. Further study is needed to quantify the difference in performance this causes.

\begin{figure}[H]
	\centering
        \includegraphics[width=\columnwidth]{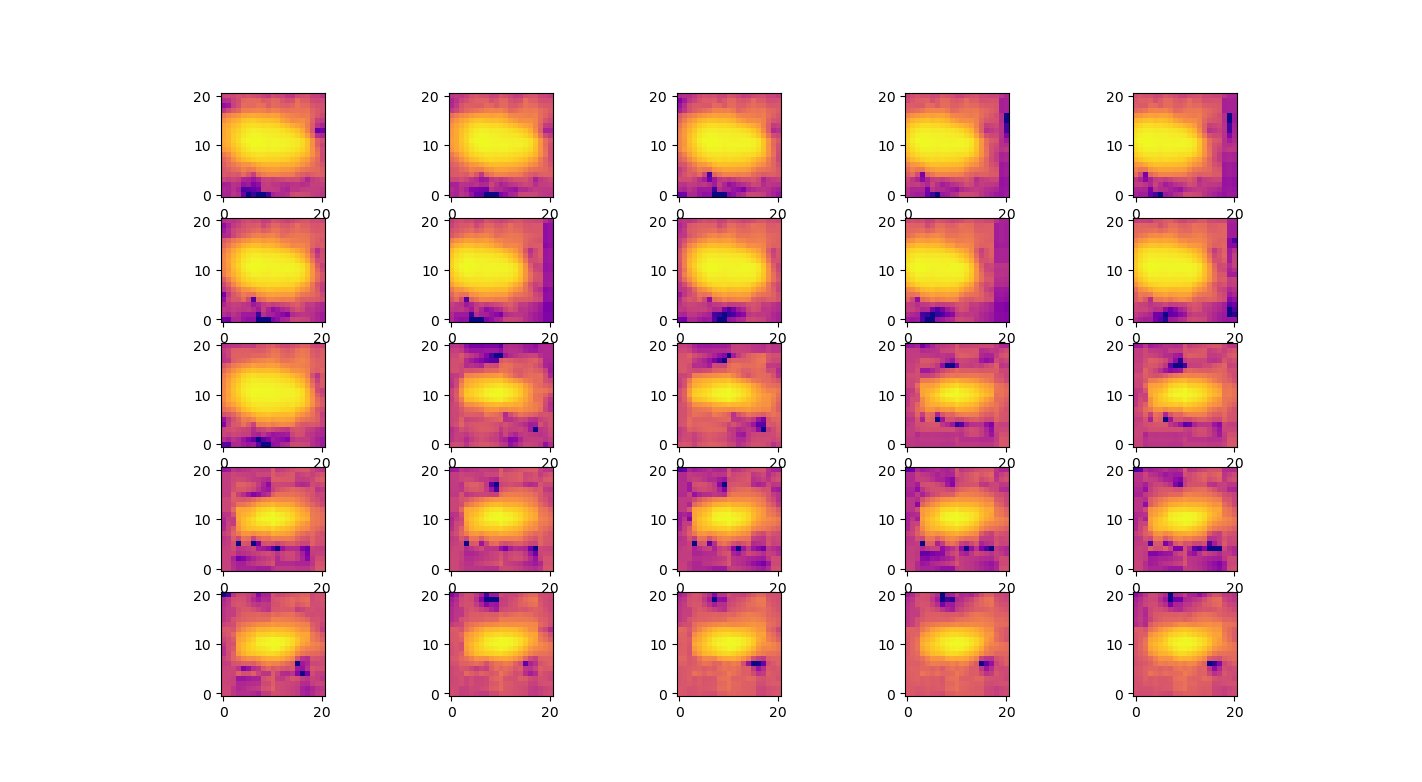}
	\caption{A sample of 25 stars collected using the optical system.}
	\label{stars}
\end{figure}

\begin{figure}[H]
	\centering
    \includegraphics[width=0.66\columnwidth]{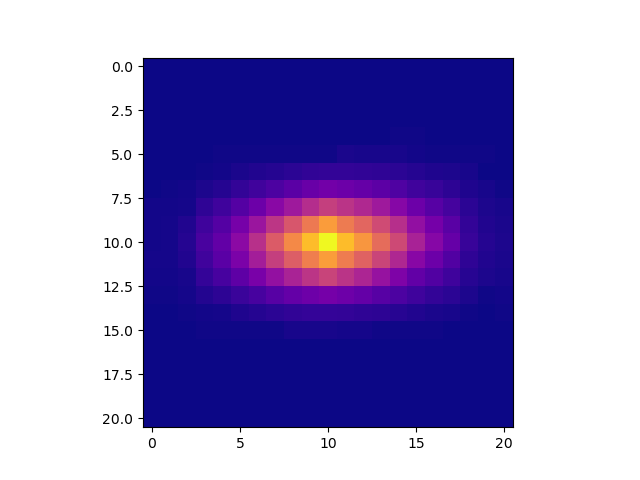}
	\caption{The effective point spread function generated from a sample of 1499 star images.}
	\label{PSF}
\end{figure}

Due to the availability of numerous images of stars with our optical system (see Fig. \ref{stars}), we use these images to construct an effective PSF (ePSF) using Anderson's method, as shown in Fig. \ref{PSF}. For a comprehensive overview, we direct readers to \cite{Anderson_2000,2016Anderson}.

 The method works by the following steps:

\begin{itemize}
    \item \textbf{Star Selection} Take a sample of bright, isolated stars to build the ePSF.
    \item  \textbf{Initial effective PSF Construction} Use the selected stars to create an initial ePSF. This involves oversampling the PSF with respect to the detector pixels to capture finer details.
    \item \textbf{Iterative Refinement} Iteratively refine the ePSF by alternating between updating the ePSF model and re-evaluating the star positions and fluxes. This process continues until the changes in star positions are below a given tolerance.
\end{itemize}

\begin{figure}[H]
	\centering
        \includegraphics[width=0.45\columnwidth]{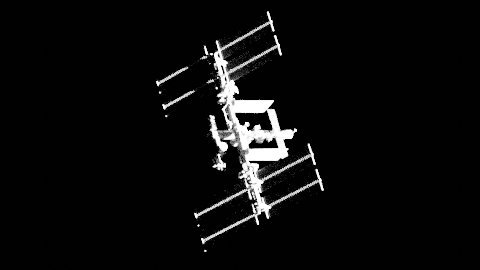}
        \includegraphics[width=0.45\columnwidth]{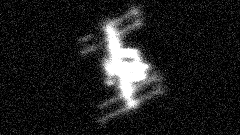}
	\caption{\textit{Left} the emulated high quality imagery, \textit{right} the emulated observed data}
	\label{EmulatedImagery}
\end{figure}

\begin{figure}[H]
	\centering
\includegraphics[width=0.45\columnwidth]{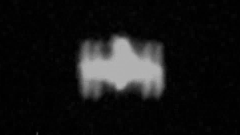}
        \includegraphics[width=0.45\columnwidth]{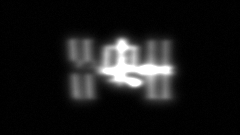}
	\caption{ \textit{Left} the actual imagery of the ISS, \textit{right} the emulated imagery of the ISS.}
	\label{RealImagery}
\end{figure}

In our application, we first convolve the point spread function over our rendered satellite images, then in addition a bloom filter add background noise - as shown in Fig. \ref{EmulatedImagery}. The effect is to create images that closely emulate those captured with the physical optical system see Fig. \ref{RealImagery}; We increased the amount of noise in the training data beyond that observed in the actual imagery. This was to make the model more general and because the level of noise varies based on photographing conditions prevalent at the time. Using this approach there is effectively no limit to how large a  synthetic dataset we can generate in order to train our models on. The only restrictions are the computation and the availability of high-quality 3D models. An additional benefit of this approach is that it can be applied to other camera systems. 

\subsection*{Deconvolution for image deblurring}
We use convolution to model the blurring observed in our RSO imagery, in the process we collect a PSF for our optical system. It is natural to investigate how we might use this effective PSF for image recovery by image deconvolution. This essentially attempts to invert the PSF convolution to recover the original image. \cite{pasha2024tripspytechniquesregularizationinverse}

\begin{itemize}
    \item $I(m,n)$ is the observed image.
    \item $O(m,n)$ is the true image of the object.
    \item $PSF(x-x',y-y')$ is the point spread function.
\end{itemize}

\begin{align}
    I[m, n] = \sum_{j=-\infty}^{\infty} \sum_{k=-\infty}^{\infty} O[j, k] \cdot PSF[m - j, n - k] \label{discreteconv}
\end{align}

We can represent the convolution of a discretized PSF over a discrete image (as shown in Eq. \eqref{discreteconv}) as a linear operator. As shown in Eq. \eqref{LinForm} we reformulate the inverse problem in terms of $A$ the forward operator (convolution of the PSF), $x$ the recovered image, $b_{true}$ the true image, and $b$ the image we observe.
\begin{align}
    Ab_{true}=b \quad \min_x ||Ax-b||_2 \label{LinForm}
\end{align}
In our case, we also need to account for an error term (Eq. \eqref{noisyLin}).
\begin{align}
    b = Ab_{true}+e \label{noisyLin}
\end{align}
In Eq, \eqref{optimFrom} we are solving for a regularized solution $x_{reg}$ by formulating the problem as an optimization problem.
\begin{align}
x_{reg} &= \arg\min_{\mathbf{x}\in \mathcal{D}\subseteq\mathbb{R}^n}\mathcal{F}(x)+\alpha\mathcal{R}(x),\quad \alpha >0 \label{optimFrom}
\end{align}
Where $\mathcal{F}$ is the fit-to-data term involving the matrix $A$ and the observed image $b$, $\alpha$ is the regularization parameter, $\mathcal{R}$ is the regularization term, $\mathcal{D}$ is a set of constraints.

Here, we use the TRIPS-PY \cite{pasha2024tripspytechniquesregularizationinverse} python package for the regularization of inverse problems. We ran the deconvolution algorithms with the following hyperparameters chosen to balance performance and quality. Arnoldi-Tikhonov, Hybrid Generalized Minimal Residual and Golub-Kahan-Tikhonov with the number of iterations as 20, 20, and 10 respectively.

\subsection*{Discrepancy principle}
To accurately recover an image using a deconvolution method we need to select a good regularization parameter. The discrepancy principle relies on prior knowledge about the norm of the noise $\delta$ introduced by the camera and seeking $\alpha$ such that the discrepancy between our convolved and deconvolved solution, which in the ideal case would be just noise (see Eq. \eqref{discrepancy}), matches the amount of noise \cite{WhitneyDiscrepancy}. All of the deconvolution methods we use employ the discrepancy principle to select the regularization term.

\begin{align}
\label{discrepancy}
    ||Ax_{reg}-b||_2 = ||Ax_{reg} - Ab_{true} - e||_2 \approx ||e||_2 = \delta \
\end{align}

\subsection*{U-Net}
\begin{figure}[H]
	\centering
        \includegraphics[width=\columnwidth]{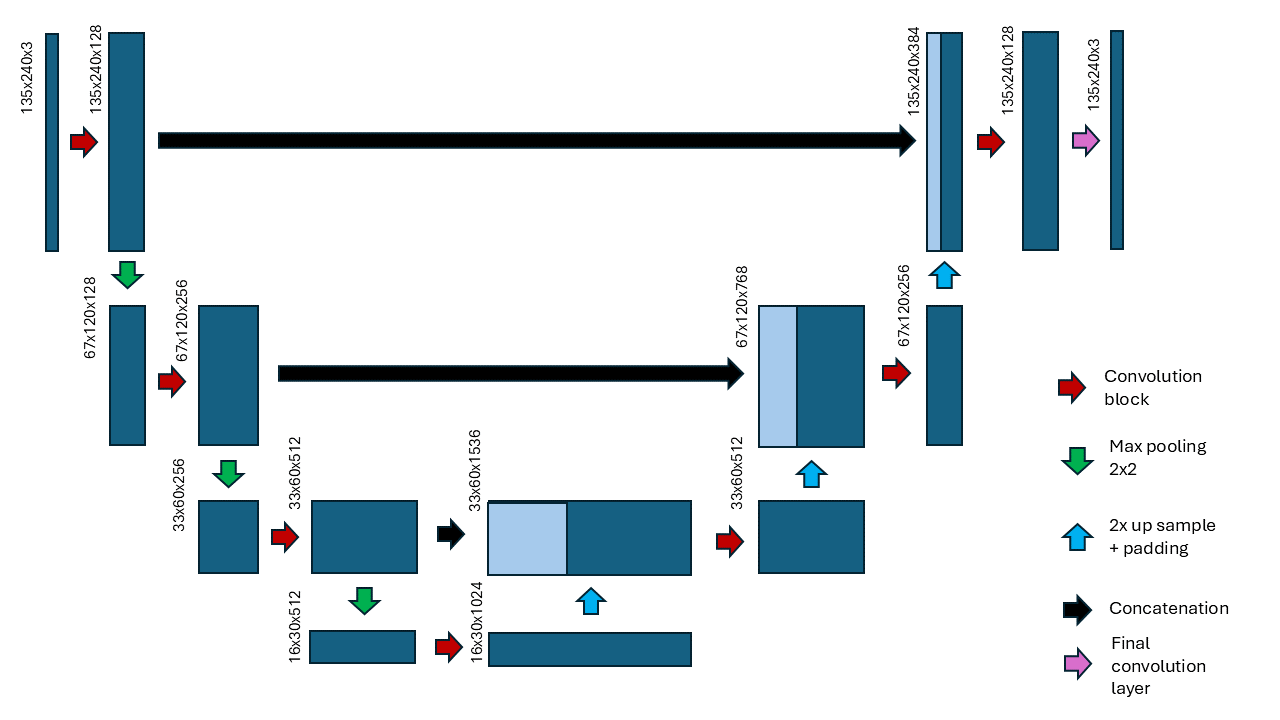}
	\caption{U-Net architecture used for image reconstruction}
	\label{U-Net}
\end{figure}

\begin{figure}[H]
	\centering
        \includegraphics[width=0.5\columnwidth]{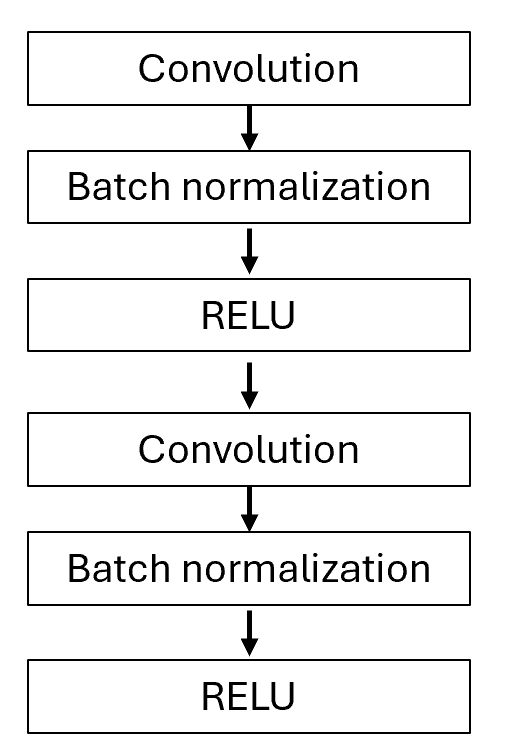}
	\caption{The convolution block comprises the majority of the U-Net model. Each consists of two sets of convolution layers with 3$\times$3 kernels, batch normalization, and a Rectified Linear Unit (ReLU) activation function}
	\label{Convolution block}
\end{figure}

We investigate the application of U-Net to image reconstruction of RSO imagery. The architecture of U-Net is a convolutional neural network with a contracting path for feature extraction and an expansive path for precise localization (see Fig. \ref{U-Net}).  The convolutional block, as shown in Fig \ref{Convolution block}, plays a crucial role within U-Net by applying filters to extract features and reduce spatial dimensions.
We have prepared a dataset of low-quality and high-quality image pairs that we used to train our model to recover the degraded image. We augment the dataset with a perspective transform, use the Adam optimizer with lr = 0.001, and look ahead with 6 steps. We use a train/test/validation split of 80/10/10. We modified the upscale steps in the network to accommodate the fact that the input dimension would be 135x240, this is the 8x downscale of 1080x1920. Because the height dimension was not even, padding the upsampling layers using wrap-around padding was necessary to have the concatenation of output layers work correctly.

\subsection*{Pose estimation}

\begin{figure}[H]
	\centering
        \includegraphics[width=0.5\columnwidth]{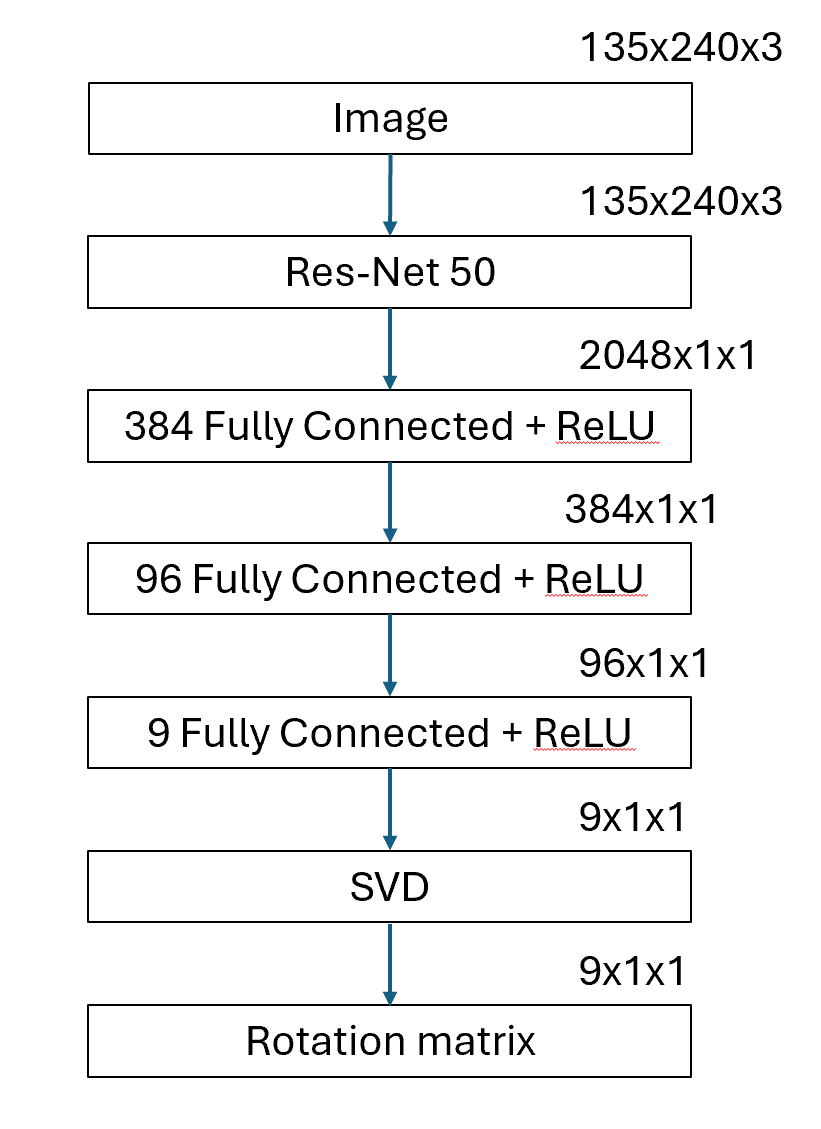}
	\caption{Model architecture used for rotation estimation}
	\label{ResNet}
\end{figure}

We use the ResNet50 model \cite{he2015deepresiduallearningimage} trained on the ImageNet \cite{5206848} for feature extraction and trained a comparatively small fully connected (FC) network of 384 and 96 nodes and 9 output nodes as shown in Fig. \ref{ResNet}. These 9 nodes represent a rotation matrix which combined with a Singular Value Decomposition had the best results. This is because SVD orthogonalization maximizes the likelihood and minimizes the expected error in the presence of Gaussian noise \cite{levinson2020analysissvddeeprotation}. Intuitively this makes sense as SVD essentially decomposes a matrix into three matrices representing rotation, scaling, and another rotation. We use the Adam optimizer with a look-ahead mechanism, a step size of 6, and a learning rate of 0.0001

\section*{Results}
\subsection*{Deconvolution}
\begin{figure}[H]
	\centering
        \includegraphics[width=0.45\columnwidth]{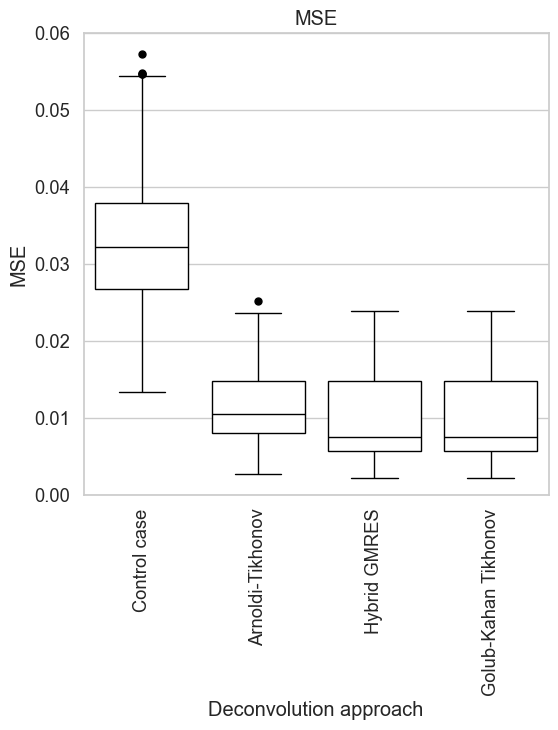}
        \includegraphics[width=0.45\columnwidth]{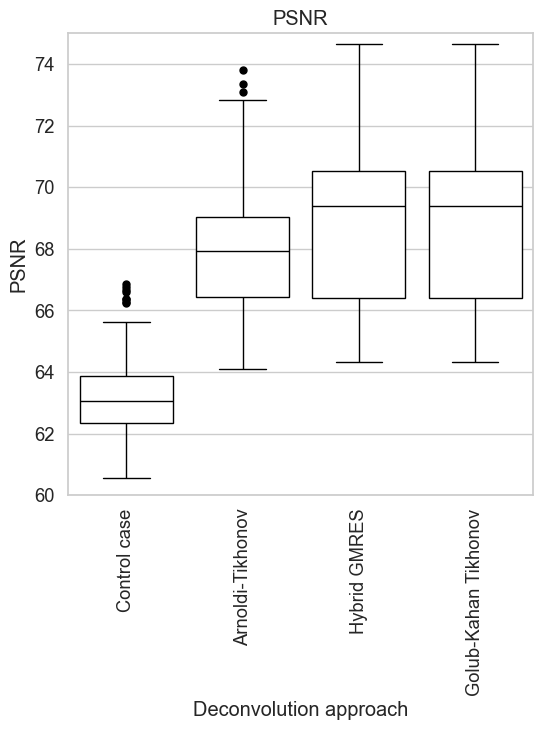}
        \includegraphics[width=0.45\columnwidth]{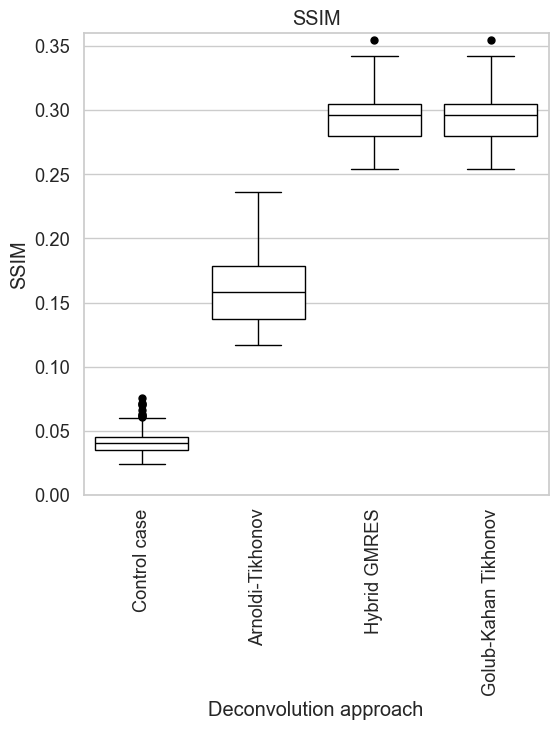}
	\caption{\textit{top left} A box plot comparison of the Mean Squared Error of deconvolution algorithms; \textit{top right} A box plot comparison of the peak Signal to Noise Ratio of deconvolution algorithms; \textit{bottom} A box plot comparison of the Structural Similarity Index Measure of deconvolution algorithms}
	\label{DeconvResults}
\end{figure}

To measure the performance of the image recovery we use MSE (Mean Squared Error), SSIM \cite{1284395} (Structural Similarity Index Error), and PSNR (Peak Signal to Noise Ratio)- all common metrics for image recovery tasks. 

First, we compare the performance of the deconvolution techniques applied to the synthetic dataset, and include the control of no deconvolution being applied to the image. Results are show in Fig. \ref{DeconvResults}. We see that these techniques improve the quality of the imagery. Arnoldi-Tikhonov performed the worst of the approaches, which could be expected as it is the most "naive" approach. Hybrid Generalised Minimal Residual and Golub-Kahan Tikhonov performed effectively identically, this is likely because both are projection approaches using the same $\delta$ so the level of regularization should be identical. More comparisons could be made concerning the stability of the algorithms and the computational cost. In the following analysis we chose to only consider Golub-Kahan Tikhonov to simplify the comparison of results.

\begin{figure}[H]
	\centering
        \includegraphics[width=0.45\columnwidth]{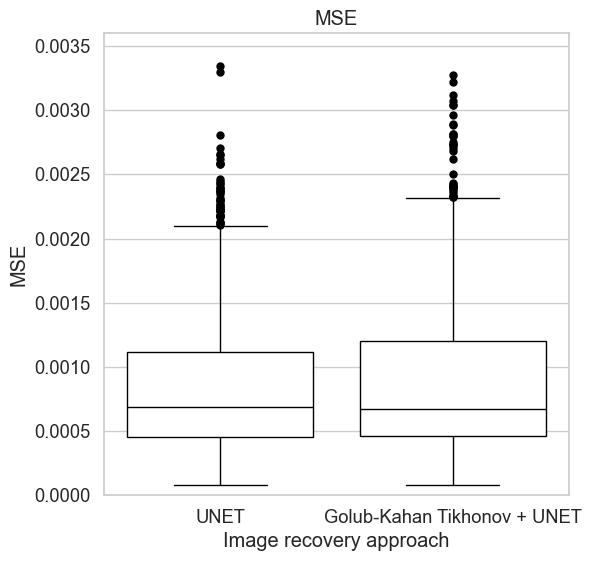}
        \includegraphics[width=0.45\columnwidth]{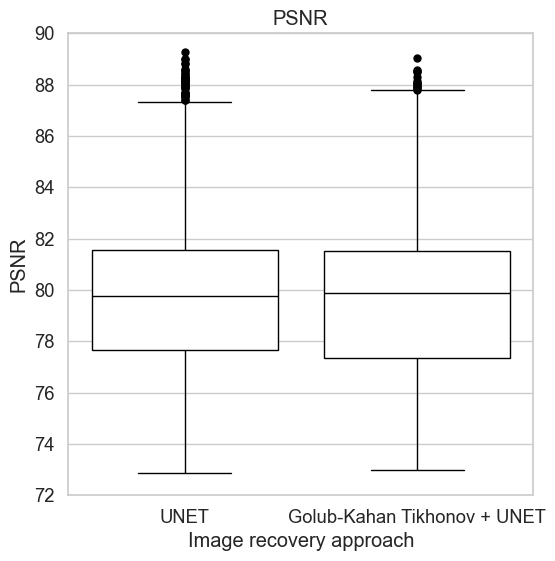}
        \includegraphics[width=0.45\columnwidth]{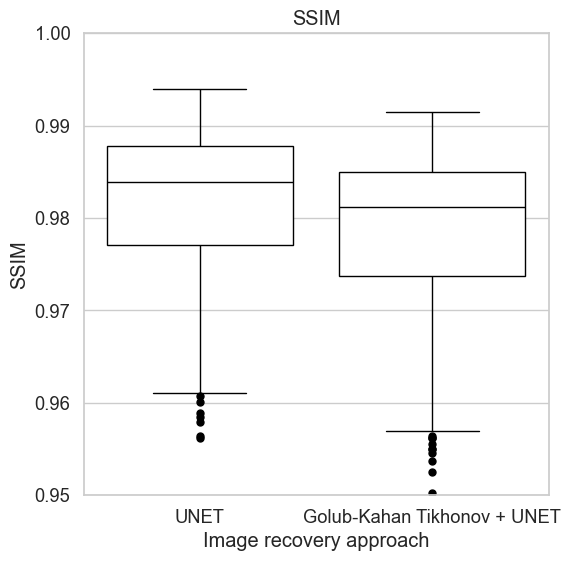}
	\caption{\textit{top left} A box plot comparison of the Mean Squared Error of image recovery approaches; \textit{top right} A box plot comparison of the peak Signal to Noise Ratio of image recovery approaches; \textit{bottom} A box plot comparison of the Structural Similarity Index Measure of image recovery approaches}
	\label{ImageRecoveryResults}
\end{figure}

\begin{table}[h]
    \centering
    \caption{Performance metrics of image reconstruction methods}
    \label{table_1}
    \begin{tabular}{l|S[table-format=1.6]|S[table-format=1.6]|S[table-format=1.6]}
        \hline
        \textbf{Image} & {\textbf{MSE}} & {\textbf{PSNR}} & {\textbf{SSIM}} \\
        \textbf{reconstruction} & & & \\
        \hline
        U-Net & 8.36 e-4 & 79.90 & 0.982 \\
        GK-Tikhonov + U-Net & 8.72 e-4 & 79.80 & 0.979 \\
        \hline
    \end{tabular}
\end{table}

\begin{figure}[H]
	\centering
        \includegraphics[width=\columnwidth]{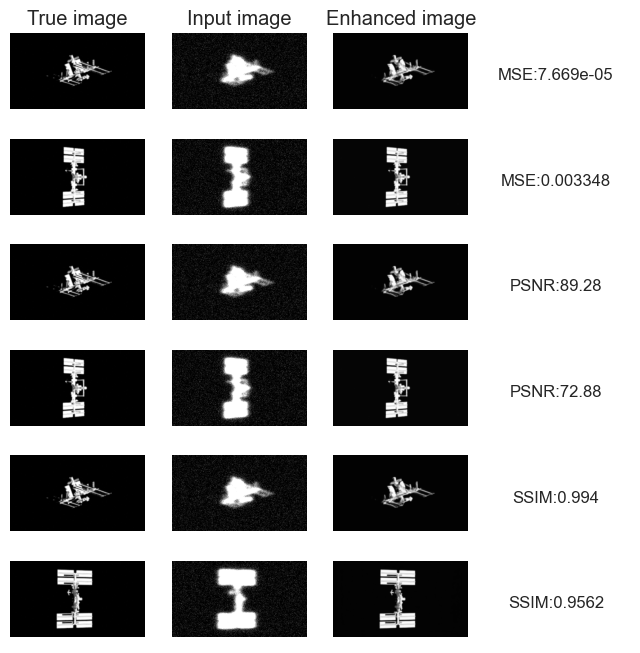}

	\caption{Demonstration of U-Net applied to simulated RSO imagery. We show the best and worst performing image for each metric (Mean Squared Error, Peak-Signal to Noise Ratio, and Structural Similarity Index Measure).}
	\label{RecoveredSIMImagery1}
\end{figure}

\begin{figure}[H]
	\centering
        \includegraphics[width=\columnwidth]{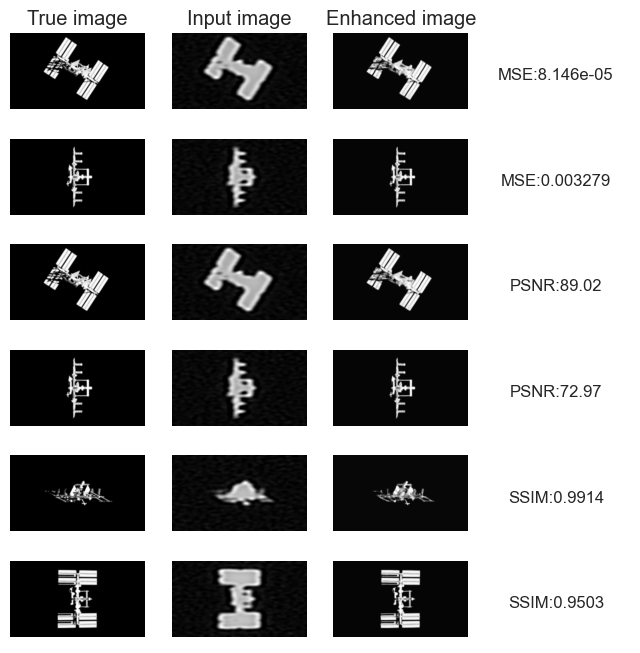}

	\caption{Demonstration of U-Net applied to simulated RSO imagery with Golub-Kahan Tikhonov deconvolution applied beforehand. We show the best and worst performing image for each metric (Mean Squared Error, Peak-Signal to Noise Ratio, and Structural Similarity Index Measure).}
	\label{RecoveredSIMImagery2}
\end{figure}

Next, we compare the performance of training a U-Net to the direct task of image recovery and use deconvolution as a pre-processing step. The results in Fig. \ref{ImageRecoveryResults} show that when a U-Net is trained directly for image recovery it outperforms combining deconvolution with U-Net. Although unexpected, this could be because in the process of deconvolving the image some important information is lost that would otherwise aid image reconstruction. In Fig. \ref{RecoveredSIMImagery1} and Fig. \ref{RecoveredSIMImagery2} we show examples of the image reconstruction performed by our models on the synthetic images. We include the best and worst example for each of the recovery metrics.
\begin{figure}[H]
	\centering
        \includegraphics[width=0.45\columnwidth]{figures/sliced_ISS.png}
        \includegraphics[width=0.45\columnwidth]{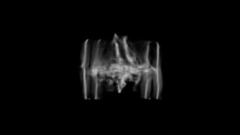}
	\caption{Demonststion of U-Net applied to real RSO imagery \textit{left} The real RSO imagery \textit{right} The recovered imagery.}
	\label{RecoveredRSOImagery}
\end{figure}

We represent the results of our rotation estimation using violin plots. A violin plot is a combination of a density plot and a box plot. The width of the violin shape represents the density of a particular result, inside the violin shape contains the quartile values.

\begin{figure}[H]
	\centering
        \includegraphics[width=0.75\columnwidth]{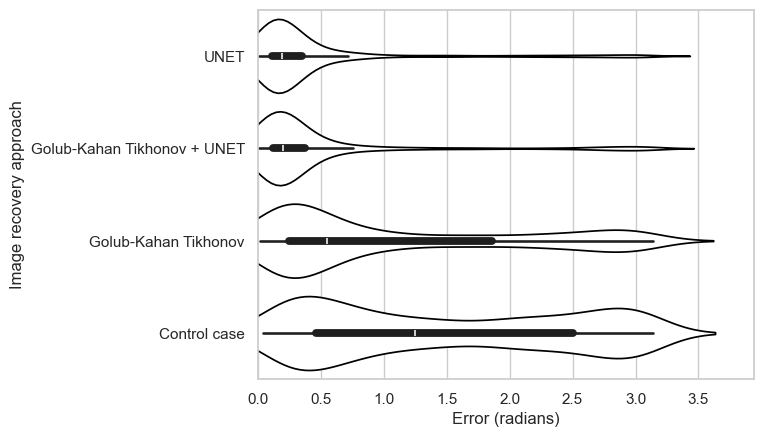}
	\caption{Violin plots demonstrating the results of our rotation estimation model when trained on datasets using different image recovery approaches.}
	\label{RotationResults}
\end{figure}

\begin{table}[h]
    \centering
    \caption{Error (radians) of rotation estimation trained with different image reconstruction methods}
    \label{table_2}
    \begin{tabular}{l|S[table-format=1.3]|S[table-format=1.3]}
        \hline
        \textbf{Image reconstruction} & {\textbf{Mean}} & {\textbf{Std. dev.}} \\
        \hline
        Control (no recovery) & 1.460 & 1.040 \\
        Golub-Kahan Tikhonov & 1.050 & 1.010 \\
        GK-Tikhonov + U-Net & 0.447 & 0.693 \\
        U-Net & 0.414 & 0.637 \\
        \hline
    \end{tabular}
\end{table}

\begin{table}[t]
    \centering
    \caption{Mean Squared Error of rotation estimation models trained with different image reconstruction methods}
    \label{table_3}
    \begin{tabular}{l|S[table-format=1.4]|S[table-format=1.3]}
        \hline
        \textbf{Image reconstruction} & {\textbf{Mean}} & {\textbf{Std. dev.}} \\
        \hline
        Control (no recovery) & 0.4020 & 0.352 \\
        Golub-Kahan Tikhonov & 0.2720 & 0.337 \\
        GK-Tikhonov + U-Net & 0.0868 & 0.217 \\
        U-Net & 0.0784 & 0.202 \\
        \hline
    \end{tabular}
\end{table}

The results in Fig. \ref{RotationResults} show that image reconstruction significantly improves the performance of our rotation estimation. The application of deconvolution to images reduced the mean error by $27.92\%$. This is a promising result and could suggest carrying over to real RSO imagery. More research is needed. The U-Net performed best on its own with only a mean error of $0.414$ compared to $0.447$ when combined with the Golub-Kahan-Tikhonov deconvolution. Our analysis reveals that, while image reconstruction generally improves model performance across the dataset, significant challenges persist at the extremes of the error distribution. The violin plot demonstrates maximum errors can reach $\pi$ radians, even in models employing reconstruction techniques. One cause could be the regularization techniques employed during image reconstruction, while effective at noise reduction, may inadvertently eliminate crucial pose information. This suggests a fundamental trade-off between noise suppression and information preservation in our pre-processing pipeline. Another could be the architectural decision to represent rotations using Euler angles, which introduces inherent mathematical limitations. The phenomenon of gimbal lock, which occurs when two rotation axes align, creates fundamental ambiguities in pose representation. This limitation is particularly relevant for poses approaching the extremes of the possible rotation space. Both quaternions and rotation matrices offer viable alternatives that avoid the gimbal lock problem entirely, potentially improving the accuracy of pose estimation at the distribution extremes.

\section*{Conclusion}
Our research examined approaches for both restoring images of RSOs and also determining their pose using direct and deep learning techniques. We introduce an innovative approach for generating realistic synthetic RSO imagery by generating an effective PSF using stars collected with a real optical system and convolving over images rendered in Blender. We demonstrate how deconvolution can be used to recover images of RSOs so that the rotation can be more accurately estimated. Additionally, we found that the U-Net architecture achieved a high level of performance in image recovery tasks, and performed the best without deconvolution as a pre-processing step. These results extend our understanding of ill-posed inverse problems and their applications to RSO imagery. This work has important implications for the development of future image recovery models. We show how a limited amount of real imagery collected with an optical system can generate a source data set that transfers to the target data set. This greatly expands the potential avenues for developing deep-learning models for space domain awareness, as the cost and difficulties involved in collecting large labeled datasets of RSOs can be prohibitive to many SDA providers. A limitation of this study is that we only considered one potential object, namely the International Space Station, so it is still to be seen how well this approach transfers to other RSOs. However, there is no reason to think that the dataset cannot be readily extended to include other RSOs for example the Tiangong space station or the Hubble telescope. Additional research is needed to see how well this approach generalizes to the diverse domain of RSO imagery and how reliably insights can be drawn from reconstructed images. Space Domain Awareness is a growing issue as more satellites enter the already congested Low-Earth orbit. The capability to accurately characterize objects for collision avoidance or debris removal is crucial to maintain a clear and cooperative space environment.

\printbibliography
\end{document}